\documentclass[mathematics,article,accept,pdftex,moreauthors]{Definitions/mdpi} 

\usepackage{booktabs}
\usepackage{amsmath,graphicx}
\usepackage{xcolor}
\usepackage[lined,boxed,ruled,commentsnumbered]{algorithm2e}

\SetCommentSty{mycommfont}

\SetKwInput{KwInput}{Input}                % Set the Input
\SetKwInput{KwOutput}{Output}              % set the Output

\newcommand{\mycomment}[1]{}

\makeatletter
\def\UrlAlphabet{%
\do\a\do\b\do\c\do\d\do\e\do\f\do\g\do\h\do\i\do\j%
\do\k\do\l\do\m\do\n\do\o\do\p\do\q\do\r\do\s\do\t%
\do\u\do\v\do\w\do\x\do\y\do\z\do\A\do\B\do\C\do\D%
\do\E\do\F\do\G\do\H\do\I\do\J\do\K\do\L\do\M\do\N%
\do\O\do\P\do\Q\do\R\do\S\do\T\do\U\do\V\do\W\do\X%
\do\Y\do\Z}
\def\UrlDigits{\do\1\do\2\do\3\do\4\do\5\do\6\do\7\do\8\do\9\do\0}
\g@addto@macro{\UrlBreaks}{\UrlOrds}
\g@addto@macro{\UrlBreaks}{\UrlAlphabet}
\g@addto@macro{\UrlBreaks}{\UrlDigits}
\makeatother

\firstpage{1} 
\pubvolume{11}
\issuenum{11}
\articlenumber{2548}
\pubyear{2023}
\copyrightyear{2023}
\externaleditor{Academic Editor: Chengjie Sun }
\datereceived{1 May 2023} 
\daterevised{26 May 2023} % Comment out if no revised date
\dateaccepted{29 May 2023} 
\datepublished{1 June 2023} 
\hreflink{https://doi.org/10.3390/\linebreak math11112548} % If needed use \linebreak
\Title{Multilingual Multiword Expression Identification Using Lateral Inhibition and Domain Adaptation}

\TitleCitation{Multilingual Multiword Expression Identification Using Lateral Inhibition and Domain Adaptation}

\Author{Andrei-Marius Avram $^{1,}$*\orcidA{}, Verginica Barbu Mititelu $^{2}$\orcidB{}, Vasile Păiș $^{2}$\orcidE{}, Dumitru-Clementin Cercel $^{1,}$*\orcidC{}\linebreak and Ștefan Trăușan-Matu $^{1,2}$\orcidD{}}

\AuthorNames{Andrei-Marius Avram, Verginica Barbu Mititelu, Vasile Păiș, Dumitru-Clementin Cercel and Ștefan Trăușan-Matu}

\AuthorCitation{Avram, A.-M.; Mititelu, V.B.; Păiș, V.; Cercel, D.-C.; Trăușan-Matu, Ș.}
\address{%
$^{1}$ \quad Computer Science and Engineering Department, Faculty of Automatic Control and Computers,
University Politehnica of Bucharest, 060042 Bucharest, Romania\\
$^{2}$ \quad Research Institute for Artificial Intelligence “Mihai Drăgănescu”, Romanian Academy,\linebreak 050711 Bucharest, Romania}

\corres{Correspondence: 
 andrei\_marius.avram@stud.acs.upb.ro (A.-M.A.);  dumitru.cercel@upb.ro (D.-C.C.)}

\abstract{Correctly identifying multiword expressions (MWEs) is an important task for most natural language processing systems since their misidentification can result in ambiguity and misunderstanding of the underlying text. In this work, we evaluate the performance of the mBERT model for MWE identification in a multilingual context by training it on all 14 languages available in version 1.2  of the PARSEME corpus. We also incorporate lateral inhibition and language adversarial training into our methodology to create language-independent embeddings and improve its capabilities in identifying multiword expressions. The evaluation of our models shows that the approach employed in this work achieves better results compared to the best system of the PARSEME 1.2 competition, MTLB-STRUCT, on 11 out of 14 languages for global MWE identification and on 12 out of 14 languages for unseen MWE identification. Additionally, averaged across all languages, our best approach outperforms the MTLB-STRUCT system by 1.23\% on global MWE identification and by 4.73\% on unseen global \mbox{MWE identification.}}


\begin{document}

\section{Introduction} \label{sec:intro}

Natural language processing (NLP) is a significant domain of artificial intelligence, with applications ranging from language translation to text classification and information retrieval. NLP allows computers to interpret and process human language, enabling them to perform tasks such as understanding and responding to questions, summarizing texts, and detecting sentiments. Some phenomena present in language can preclude its correct understanding by machines (and even humans sometimes). Such a phenomenon is represented by multiword expressions (MWEs), which are groups of words that function as a unit and convey a specific meaning that is not the sum of the meanings of the component words (i.e., the expression lacks compositionality). Examples of MWEs include idioms (e.g., ``break a leg'' is used to wish someone good luck), collocations (e.g., ``take an exam''), or compounds (e.g., ``ice cream''), different authors assuming a more comprehensive or a narrower meaning of this term. The number of MWEs in a language is relatively high.
The authors of \cite{Shudoetal2011} synthesized papers reporting the number or proportion of MWEs in different languages: English---with an almost equal number of MWEs and single words; French---with  3.3 times greater number of MWE adverbs than that of single adverbs and 1.7~times greater number of MWE verbs than that of single verbs; and Japanese---in which 44\% of the verbs are MWEs. Despite being so numerous in the dictionary, MWEs' frequency in corpora is low \cite{Savary2008}. 

Identifying and processing MWEs is crucial for various NLP tasks \cite{avram2023romanian}. In machine translation, for instance, the correct translation of an MWE often depends on the specific context in which it appears. Suppose an MWE is translated rather than appropriately localized for the target language. In that case, the resulting translation may be difficult to understand for native speakers or may convey a wrong meaning \cite{zaninello2020multiword}. In text classification tasks, MWEs are considered essential clues regarding the sentiment or topic of a text \cite{najar2018multi}. Additionally, to improve the accuracy of search engines in information retrieval, MWEs can help disambiguate the meaning of a query \cite{goyal2020development}.

%Vergi: 2-3 paragrafe despre PARSEME:MWE
Acknowledged recent progress in the field has been made by the PARSEME community \cite{Savary-et-al:2018}, which evolved from the COST action with the same name, where the topics of interest were parsing and MWEs
(\url{https://typo.uni-konstanz.de/parseme/} last accessed on 21 April 2023). There are two significant outcomes of their activity, (i) a multilingual corpus annotated for verbal MWEs (VMWEs) in 26 languages by more than 160 native annotators, with three versions so far (\url{https://lindat.mff.cuni.cz/repository/xmlui/handle/11372/LRT-2282}, \url{https://lindat.mff.cuni.cz/repository/xmlui/handle/11372/LRT-2842}, \url{https://lindat.mff.cuni.cz/repository/xmlui/handle/11234/1-3367} last accessed on 21 April 2023) \cite{corpus-1.0,corpus-1.1,corpus-1.2}; and (ii) a series of shared tasks (also three editions so far) dedicated to the automatic and semi-supervised identification of VMWEs in \mbox{texts \cite{shared-task-1.0,shared-task-1.1,shared-task-1.2}}, in which the previously mentioned corpora were used for training and testing the participating systems.

Developing systems that can handle multiple languages is another important NLP area. In particular, the ability to accurately process and analyze text in various languages is becoming increasingly important as the world becomes more globalized and interconnected. For example, multilingual NLP systems can improve machine translation, allowing computers to translate text from one language to another accurately. This can be particularly useful in situations where there is a need to communicate with speakers of different languages, such as in global business or international relations. In addition to its practical applications, multilingual NLP is an important area of study from a theoretical perspective. Research in this field can help shed light on the underlying principles of language processing and how these principles differ across languages \cite{ponti2019modeling,arroyo2021learning}.

Multilingual Transformer models have become a popular choice for multilingual NLP tasks due to their ability to handle multiple languages and achieve strong performance on a wide range of tasks. Based on the Transformer architecture \cite{vaswani2017attention}, these models are pre-trained on large amounts of multilingual data and can be fine-tuned for specific NLP tasks, such as language translation or text classification. Some models that have become influential in this area include the multilingual bidirectional encoder from transformers (mBERT) \cite{devlin2019bert}, cross-lingual language model (XLM) \cite{conneau2019cross},  XLM-RoBERTa (XLM-R) \cite{conneau2020unsupervised}, and multilingual bidirectional auto-regressive transformers (mBART) \cite{liu2020multilingual}. One of the essential benefits of multilingual Transformer models is their ability to transfer knowledge between languages. These models can learn common representations of different languages, allowing them to perform well on tasks in languages that they have yet to be specifically trained on. Thus,  multilingual Transformer models are a good choice for NLP tasks that involve multiple languages, such as machine translation or cross-lingual information retrieval \cite{kalyan2021ammus}.

In this work, we leverage the knowledge developed in the two research areas (i.e., MWEs and multilingual NLP) to improve the results obtained at the PARSEME 1.2 shared task \cite{shared-task-1.2}. We explore the benefits of combining them in a singular system by jointly fine-tuning the mBERT model on all languages simultaneously and evaluating it separately. In addition, we try to improve the performance of the overall system by employing two mechanisms, (i) the newly introduced lateral inhibition layer \cite{pais-2022-racai} on top of the language model and  (ii) adversarial training \cite{lowd2005adversarial} between languages. For the last mechanism, other researchers have experimented with this algorithm and have shown that it can provide better results with the right setting \cite{dong2020leveraging}; however, we are the first to experiment with and show the advantages of lateral inhibition in multilingual adversarial training.

Our results demonstrate that by employing lateral inhibition and multilingual adversarial training, we improve the results obtained by MTLB-STRUCT \cite{taslimipoor2020mtlb}, the best system in edition 1.2 of the PARSEME competition, on 11 out of 14 languages for global MWE identification and 12 out of 14 languages for unseen MWE identification. Furthermore, averaged across all languages, our highest-performing methodology achieves F1-scores of 71.37\% and 43.26\% for global and unseen MWE identification, respectively. Thus, we obtain an improvement of 1.23\% for the former category and a gain of 4.73\% for the latter category compared to the MTLB-STRUCT system.    

The rest of the paper is structured as follows. Section \ref{sec:rel_work} summarises the contributions of the PARSEME 1.2 competition and the main multilingual Transformer models.
The following section, Section \ref{sec:methodology}, outlines the methodology employed in this work, including data representation, lateral inhibition, adversarial training, and how they were employed in our system. \mbox{Section \ref{sec:eval_setup}} describes the setup (i.e., dataset and training parameters) used to evaluate our models. Section \ref{sec:res} presents the results, and Section \ref{sec:discussion} details our interpretation of their significance. Finally, our work is concluded in Section \ref{sec:concl} with potential future \mbox{research directions.}

\section{Related Work}
\label{sec:rel_work}

\subsection{Multilingual Transformers}
This subsection will present the most influential three multilingual language models (MLLMs): mBERT, XLM, and XLM-R.
The mBERT model, similar to the original BERT model \cite{devlin2019bert}, is a Transformer model \cite{vaswani2017attention} with 12 hidden layers. However, while BERT was trained solely on monolingual English data with an English-specific vocabulary, mBERT is trained on the Wikipedia pages of 104 languages and uses a shared word-piece vocabulary. mBERT has no explicit markers indicating the input language and no mechanism specifically designed to encourage translation-equivalent pairs to have similar representations within the model. Although simple in its architecture, due to its multilingual representations, mBERT's robustness to generalize across languages is often surprising, despite needing to be explicitly trained for cross-lingual generalization. The central hypothesis is that using word pieces common to all languages, which must be mapped to a shared space, may lead to other co-occurring word pieces being mapped to this shared space \cite{pires2019multilingual}.

XLM resulted from various investigations made by the authors in cross-lingual pre-training. They introduce the translation language modeling objective (TLM), which extends the masked language modeling (MLM) objective to pairs of parallel sentences. The reason for doing that is sound and straightforward. Suppose the model needs to predict a masked word within a sentence from a given language. In that case, it can consider that sentence and its translation into a different language. Thus, the model is motivated to align the representations of both languages in a shared space. Using this approach, XLM obtained state-of-the-art (SOTA) results on supervised and unsupervised machine translation using the WMT'16 German--English and WMT'16 Romanian--English datasets \cite{bojar2016results}, respectively. In addition, the model also obtained SOTA results on the Cross-lingual Natural Language Inference (XNLI) corpus \cite{conneau2018xnli}.

\textls[20]{In contrast to XLM, XLM-R does not use the TLM objective and instead trains \mbox{RoBERTa \cite{liu2019roberta}}} on a large, multilingual dataset extracted from CommonCrawl  (\url{http://commoncrawl.org/} last accessed on 21 April 2023) datasets. In 100 languages, totaling 2.5 TB of text. It is trained using only the MLM objective, similar to RoBERTa, the main difference between the two being the vocabulary size, with XLM-R using 250,000 tokens compared to RoBERTa's 50,000 tokens. Therefore, XLM-R is significantly larger, with \mbox{550 million} parameters, compared to RoBERTa's 355 million parameters. The main distinction between XLM and XLM-R is that XLM-R is fully self-supervised, whereas XLM requires parallel examples that may be difficult to obtain in large quantities. In addition, this work demonstrated for the first time that it is possible to develop multilingual models that do not compromise performance in individual languages. XLM-R obtained similar results to monolingual models on the GLUE \cite{wang2018glue} and XNLI benchmarks.
\subsection{PARSEME 1.2 Competition}

We present the results obtained by the systems participating in edition 1.2 of the PARSEME shared task \cite{shared-task-1.2} on discovering VMWEs that were not present (i.e., were not seen) in the training corpus. We will not focus on the previous editions of this shared task for two reasons, (i) the corpora were different, on the one hand, concerning the distribution of seen and unseen VMWEs in the train/dev/test sets, and, on the other hand, smaller for some languages; and (ii) the focus in the last edition, unlike the first two, was on the systems' ability to identify VMWEs unseen in the train and dev corpora, exploring alternative ways of discovering them. Thus, in a supervised machine learning approach, the systems were supposed to learn some characteristics of seen VMWEs and, based on those, find others in the test dataset.

The competing systems used recurrent neural networks \cite{yirmibesoglu-gungor-2020-ermi,gombert2020multivitaminbooster,taslimipoor2020mtlb,kurfali2020travis}, but also exploited the syntactic annotation of the corpus \cite{pasquer2020seen2unseen,colson2020hmsid}, or association measures \cite{colson2020hmsid,pasquer2020seen2unseen}. 
The shared task was organized on two tracks, closed and open.  The former allowed only for the use of the train and dev sets provided by the organizers, as well as of the raw corpora provided for each language, with sizes between 12 and 2474 million tokens. The latter track allowed for the use of any existing resource for training the system, and examples of such resources are as follows, VMWEs lexicons in the target language or another language (exploited due to their translation in the target language) or language models (monolingual or multilingual BERT \cite{kurfali2020travis,taslimipoor2020mtlb}, XLM-RoBERTa \cite{gombert2020multivitaminbooster}). Only two systems participated in the closed track, while seven participated in the open one.

The best-performing system in the open track is MTLB-STRUCT \cite{taslimipoor2020mtlb}. It is a neural language model relying on pre-trained multilingual BERT and learning both MWEs and syntactic dependency parsing, using a tree CRF network \cite{ruch-crf}. The authors explain that the joint training of the tree CRF and a Transformer-based MWE detection system improves the results for many languages.

\textls[-20]{The second and third place in the same track is occupied by the model called \mbox{TRAVIS \cite{kurfali2020travis}}} that came in two variants, TRAVISmulti (ranked second), which employs multilingual contextual embeddings, and TRAVISmono (ranked third), which employs monolingual ones. These systems rely solely on embeddings, and no other feature is used. The author claims that the monolingual contextual embeddings are much better at generalizations than the multilingual ones, especially concerning unseen MWEs.

\section{Methodology}
\label{sec:methodology}

\begin{figure}[H]
   \hspace{-.5cm} \includegraphics[width=\textwidth]{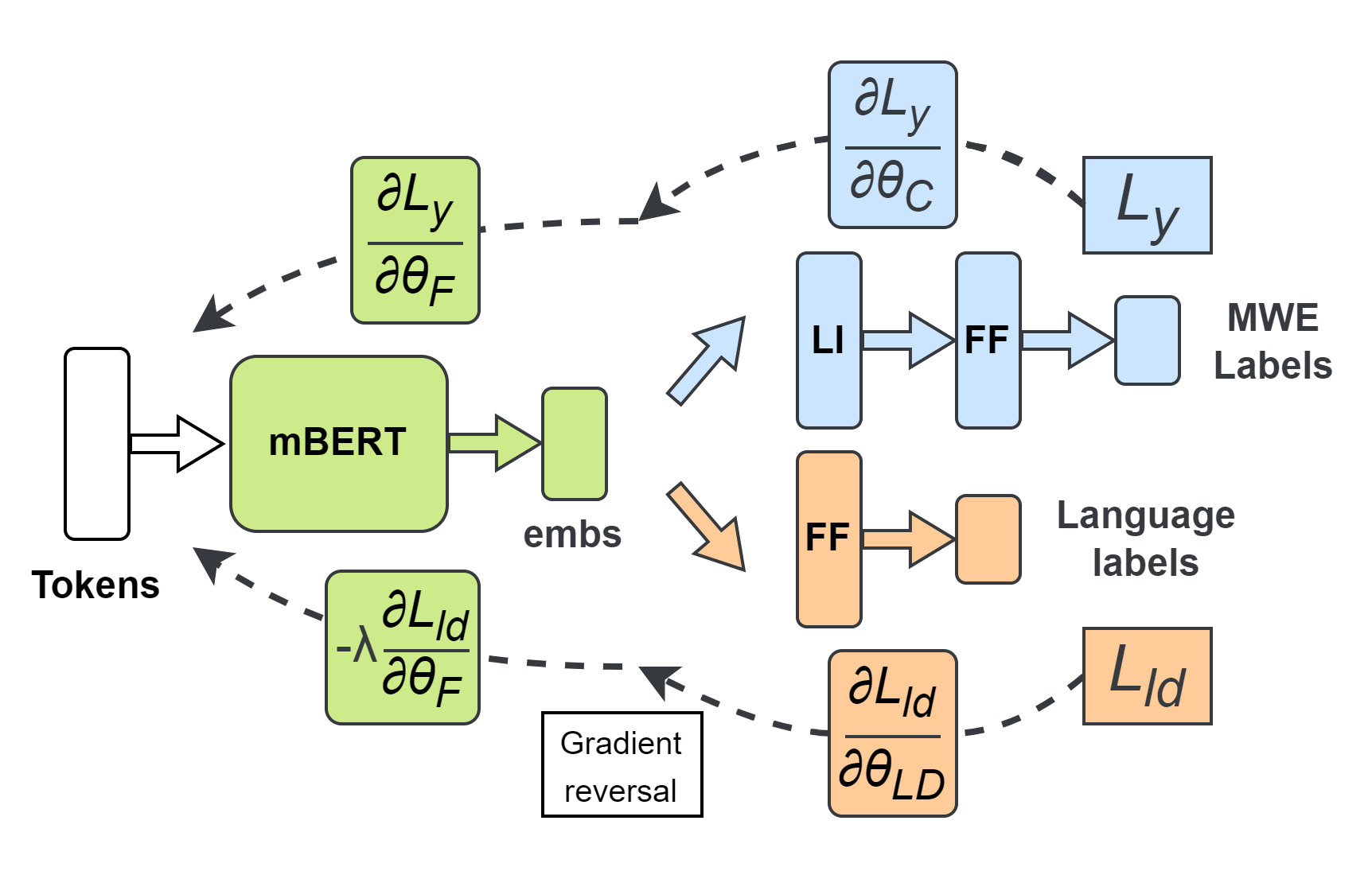}
    \caption{Domain adversarial training algorithm. We have the mBERT feature extractor $F$ with green, whose role is to generate the token embeddings, the MWE label classifier $C$ with blue, and the language classifier $LD$ with orange, whose gradient is reversed and scaled by $\lambda$ before it is fed into the feature extractor. Additionally, $C$ has incorporated in its architecture the lateral inhibition~mechanism.}
    \label{fig:dann}
\end{figure}

In this work, we perform two kinds of experiments, (i) train a model using only the data for a specific language (referred to as monolingual training) and (ii) put multiple corpora from different languages in one place, train the multilingual model on it and then evaluate the trained model on the test set of each language (referred to as multilingual training). For the latter, we also perform additional experiments to improve the results by employing lateral inhibition and adversarial training mechanisms, as depicted in Figure \ref{fig:dann}.

\subsection{Data Representation}

BERT has significantly impacted the field of NLP and has achieved SOTA performance on various tasks. Its success can be attributed to the training process, which involves learning from large amounts of textual data using a Transformer model and then fine-tuning it on a smaller amount of task-specific data.~The masked language modeling objective used during pre-training allows the model to learn effective sentence representations, which can be fine-tuned for improved performance on downstream tasks with minimal task-specific training data. The success of BERT has led to the creation of language-specific versions of the model for various languages, such as CamemBERT (French) \cite{martin2020camembert}, AfriBERT \mbox{(Afrikaans) \cite{ralethe2020adaptation}}, FinBERT (Finnish) \cite{virtanen2019multilingual}, and RoBERT (Romanian) \cite{dumitrescu2020birth}.

\vspace{-15pt}

%However, it is only possible to train language-specific models for a few languages with the necessary data and resources.

The scarceness of data and resources has resulted in recent advances in NLP being limited to English and a few high-resource languages rather than being more widely applicable across languages. To address this issue, MLLMs have been developed and trained using large amounts of unlabeled textual data collected from multiple languages. These models are designed to benefit lower resource languages by leveraging their shared vocabulary, genetic relatedness, or contact relatedness with higher resource \mbox{languages \cite{doddapaneni2021primer,math10183236}}. Many different MLLMs are available, which vary in terms of their architecture, training objective, data used for pre-training, and the number of languages covered. However, in our experiments, we employ only the mBERT model because it allows us to provide a cleaner comparison with the monolingual BERT models and thus emphasizes the strengths of our approach.

\subsection{Lateral Inhibition}
\label{subsec:li}

The biological process of lateral inhibition represents the capacity of excited neurons to reduce the activity of their neighbors \cite{cohen2011lateral}. In the visual cortex, this process is associated with an increased perception under challenging environments, such as low-lighting conditions. Previously, we proposed implementing the lateral inhibition mechanism in artificial neural networks (ANN) to improve the named entity recognition task \cite{pais-2022-racai,mitrofan-pais-2022-improving}. The intuition behind introducing this mechanism is that it reduces noise associated with word representations in some instances, such as less frequent words or contexts.

The implementation uses an additional ANN layer that filters the values of a neuron from a previous layer (the word embedding representation) based on values from other adjacent neurons in the previous layer. Equation (\ref{eq:matrices}) describes the new layer's forward pass. Here, $X$ is the layer's input vector (a token embedding representation), $Diag$ is a matrix with the diagonal set to the vector given as a parameter, $ZeroDiag$ produces a matrix with the value zero on the main diagonal, and $W$ and $B$ represent the weights and bias. $\Theta$ is the Heaviside function, described in Equation (\ref{eq:heaviside}). The derivative of the Heaviside function in the backward pass is approximated with the sigmoid function using a scaling parameter \mbox{$k$ \cite{Wunderlich2021}} (see Equation (\ref{eq:sigmoid})), a method known as surrogate gradient learning \citep{8891809}. \vspace{6pt}
\begin{equation}
\label{eq:matrices}
F(X)=X * Diag(\Theta(X * ZeroDiag(W^T) + B ))
\end{equation}

\begin{equation}
\label{eq:heaviside}
\Theta(x) = \left\{
\begin{matrix}
	1, x > 0 \\
	0, x \leq 0
\end{matrix}
\right.
\end{equation}

\begin{equation}
\label{eq:sigmoid}
\sigma(x) = \frac{1}{1+e^{-kx}}
\end{equation}

\subsection{Adversarial Training}

In recent years, adversarial training of neural networks had a significant influence, particularly in computer vision, where generative unsupervised models have demonstrated the ability to generate new images \cite{gui2021review}. A crucial challenge in adversarial training is finding the proper balance between the generator and the adversarial discriminator. As a result, several methods have been proposed in recent times to stabilize the training process \cite{wiatrak2019stabilizing,math11081926,zhang2022metasid}.  Therefore, \citet{joty-etal-2017-cross} introduced cross-lingual adversarial neural networks designed to learn discriminative yet language-invariant representations. In this work, we use the same methodology to learn task-specific representations in a cross-lingual setting and improve the predictive capabilities of a multilingual BERT model.

Our approach is rooted in the Domain Adversarial Neural Network (DANN) algorithm, initially designed for domain adaptation \cite{ganin2016domain}. DANN consists of a deep feature extractor $F$, responsible for extracting relevant features $f$ from the input data, and a deep label classifier $C$, which uses those features to make predictions about the label of the input $x$. Together, these two components form a standard feed-forward architecture. In order to improve the performance of the model on a target domain where labeled data are scarce, an additional component is added to the architecture, called a domain classifier $D$, which is responsible for distinguishing between samples from the source and target domains $d$. This domain classifier is connected to the feature extractor via a gradient reversal layer, which multiplies the gradient by a negative constant during training. The gradient reversal layer helps ensure that the feature distributions over the two domains are as similar as possible, resulting in domain-invariant features that can better generalize to the target domain. The overall training process minimizes the label prediction loss on the source examples and the domain classification loss on all samples. Thus, we have the following equations that are used to update the parameters of each of the three components:
\begin{equation}
    \begin{array}{c}
        \theta_C = \theta_C - \alpha\frac{\partial L_y}{\partial \theta_C} \\
        \theta_D = \theta_D - \alpha\frac{\partial L_d}{\partial \theta_D} \\
        \theta_F = \theta_F - \alpha (\frac{\partial L_y}{\partial \theta_F} - \lambda \frac{\partial L_d}{\partial \theta_F})
    \end{array}
\end{equation}
where $\theta_C$ are the parameters of the label classifier, $L_y$ is the loss obtained by the label classifier when predicting the class labels $y$, $\theta_D$ are the parameters of the domain classifier, $L_d$ is the loss obtained by the domain classifier when predicting the domain labels $d$, $\theta_F$ are the parameters of the feature extractor, $\lambda$ is the hyperparameter used to scale the reverse gradients, and $\alpha$ is the learning rate.

%\section{System Description}
%\label{sec:sys_descr}

\subsection{Monolingual Training}

%\begin{figure}
%    \centering
%    \includegraphics[width=\textwidth]{images/li_ner.png}
%    \caption{Lateral inhibition integration into the BERT fine-tuning methodology for MWE identification.}
%    \label{fig:li_ner}
%\end{figure}

In the monolingual training experiments, we treat the MWE task as sequence tagging, so we try to predict a label for each input token. To attain that, we employ a feed-forward layer that maps the embeddings produced by a BERT model into the specific MWE class logits and then apply the softmax activation function to obtain the probabilities. This mechanism is succinctly described in the following equation:
\begin{equation}
    p_i = softmax (e_i W^T + b)
    \label{eq:mono_train}
\end{equation}
where $p_i$ are the class MWE probabilities for the token $i$, $e_i$ are the embeddings produced by the language model, $W^T$ is the transpose of the feed-forward layer, and $b$ is its bias. We use the same BERT models for each language as in \cite{taslimipoor2020mtlb}.

\subsection{Multilingual Training}

We fine-tune the mBERT model for multilingual training using the same methodology as in the monolingual case. However, we improve the predictions by first employing the lateral inhibition layer on top of the embeddings. The lateral inhibition layer has been shown to improve the performance of language models in named entity recognition tasks \cite{pais-2022-racai,mitrofan-pais-2022-improving,avram2022racai}, and we believe that it would do the same for MWE identification since the methodology is similar for the two tasks. Therefore, the equation that describes the resulting system becomes:
\begin{equation}
    p_i = softmax (LI(e_i) W^T + b)
\end{equation}
where $LI$ is the lateral inhibition layer and the rest of the terms are the same as in \mbox{Equation (\ref{eq:mono_train}).}

We also adapt the multilingual training by employing the DANN algorithm with a language discriminator instead of the domain discriminator. Thus, we create language-independent features out of the mBERT model by reversing the gradient that comes out of the language discriminator when backpropagating through the language model. The gradient reversal mechanism in our system is described using the following equations
\begin{equation}
    \begin{array}{c}
        \theta_C = \theta_C - \alpha\frac{\partial L_y}{\partial \theta_C} \\
        \theta_{LD} = \theta_{LD} - \alpha\frac{\partial L_{ld}}{\partial \theta_{LD}} \\
        \theta_F = \theta_F - \alpha (\frac{\partial L_y}{\partial \theta_F} - \lambda \frac{\partial L_{ld}}{\partial \theta_F})
    \end{array}
\end{equation}
where $\theta_C$ are the parameters of the MWE classifier, $L_y$ is the loss obtained by the MWE classifier when predicting the MWE labels $y$, $\theta_{LD}$ are the parameters of the language discriminator, $L_{ld}$ is the loss obtained by the language discriminator when predicting the language labels $ld$, $\theta_F$ are the parameters of the mBERT model (i.e., the feature extractor in DANN), $\lambda$ is the hyperparameter used to scale the reversed gradients, and $\alpha$ is the \mbox{learning rate.}

Finally, we employ the lateral inhibition layer and the DANN methodology with a language discriminator on the mBERT model for multilingual training. The forward procedure of this approach, which is used to compute the loss between the predicted MWE probabilities for a given text and the corresponding ground truths, and the loss between the predicted language probabilities and the corresponding ground truths of the given text, is described in Algorithm \ref{alg:forward} as follows:
\begin{itemize}
    \item Tokenize the $text$ using the mBERT tokenizer, obtaining the tokens $tok_i$ (Line 1).
    \item Generate the multilingual embeddings $emb_i$ for each of the above tokens $tok_i$ using the mBERT model (Line 2).
    \item Apply the lateral inhibition layer on each of the embeddings $emb_i$ (Line 3).
    \item Use the MWE classifier composed of lateral inhibition layer output to produce the probabilities $\hat{y}_i$  of a token to belong to a certain MWE class (Line 4).
    \item Use the language discriminator on the embedding $emb_{[CLS]}$ corresponding to the token \texttt{[CLS]} to produce the probabilities $\hat{ld}_i$ of the text to belong to a certain language (Line~5). %$\hat{ld}_i$ 
    \item Compute the loss $L_y$ between the predicted MWE probabilities and the ground truth MWE labels (Line 6) and the loss $L_{ld}$ between the predicted language probabilities and the ground truth language labels (Line 7).
\end{itemize}

In Algorithm \ref{alg:backward}, we outline the backward procedure used to update the parameters of our models as follows:

\begin{itemize}
    \item Compute the gradients $\nabla_C$ for the MWE classifier using the MWE  loss $L_y$ (Line 1).
    \item Compute the gradients $\nabla_{LD}$ for the language discriminator using the language discriminator loss $L_{ld}$ (Line 2).
    \item Compute the gradients $\nabla_F$ of the mBERT model  using $\nabla_C$ and $-\nabla_{LD}$ multiplied by $\lambda$ (Line 3).
    \item Update the model parameters (i.e., $\theta_C$, $\theta_{LD}$, and $\theta_F$) using the gradient descent algorithm (Lines 4-6).
\end{itemize}
\vspace{-6pt}
\begin{algorithm}
\DontPrintSemicolon
  
  \KwInput{text, ground truth MWE labels $y_i$, and ground truth language labels $ld_i$}
  \KwOutput{MWE identification loss $L_y$ and language discrimination loss $L_{ld}$}
  
  $tok_i \gets$ tokenize($text$)\;
  $emb_i \gets$ mbert($tok_i$)\;
  $h_i \gets$ lateral\_inhibition($emb_i$)\;
  $\hat{y}_i \gets$ mwe\_classifier($h_i$)\;
  $\hat{ld}_i \gets$ language\_discriminator($emb_{[CLS]}$)\;
  $L_y \gets$ cross\_entropy\_loss($y_i, \hat{y}_i$)\;
  $L_{ld} \gets$ cross\_entropy\_loss($ld_i, \hat{ld}_i$)\;

\caption{Algorithm describing the forward pass of the multilingual training with lateral inhibition and language adversarial training.}
\label{alg:forward}
\end{algorithm}

\vspace{-6pt}

\begin{algorithm}
\DontPrintSemicolon
  
  \KwInput{MWE identification loss $L_y$, language discrimination loss $L_{ld}$, and reversed gradient scaling factor $\lambda$}
  %\KwOutput{\myBlue{Updated weights for the MWE classifier $\theta_C$, language discriminator $\theta_{LD}$, and the mBERT model $\theta_F$}}
  \KwOutput{Parameters $\theta_C$, $\theta_{LD}$, and $\theta_F$}
  $\nabla_C \gets$ compute\_gradients($L_y$)\;
  $\nabla_{LD} \gets$ compute\_gradients($L_{ld}$)\;
  $\nabla_F \gets$ compute\_gradients($\nabla_C -\lambda \nabla_{LD}$)\;
  $\theta_C \gets$ update\_parameters($\nabla_C$)\;
  $\theta_{LD} \gets$ update\_parameters($\nabla_{LD}$)\;
  $\theta_F \gets$ update\_parameters($\nabla_F$)\;
  
\caption{Algorithm describing the backward pass of the multilingual training with lateral inhibition and language adversarial training.}
\label{alg:backward}
\end{algorithm}

\vspace{-9pt}
\section{Experimental Settings}
\label{sec:eval_setup}

\subsection{Dataset}

The corpus used to evaluate our models is the PARSEME dataset version 1.2. The corpus was manually annotated with VMWEs of several types. Some are universal because they exist and were annotated in all languages in the project. These universal types are verbal idioms (e.g., the Romanian ``a face din țânțar armăsar''---eng. ``to make a mountain out of a molehill'') and light verb constructions (e.g., the Romanian ``a face o vizită''---eng. ``to pay a visit'') in which their verb is light in the sense that its semantic contribution to the meaning of the whole expression is almost null, its role being rather only that of carrying the verb specific morphological information, such as tense, number, or person.
There are also light verb constructions in which the verb carries a causative meaning (e.g., the Romanian ``a da bătăi de cap''---eng. ``to give a hard time''), and they are also annotated in all languages. The types of VMWEs that apply only to some of the languages in the project are called quasi-universal: inherently reflexive verbs (e.g., the Romanian ``a-și imagina''---eng. ``to imagine (oneself)''), verb-particle constructions (e.g., ``to give up''), multi-verb constructions (e.g., ``make do''), and inherently adpositional verbs (e.g., ``to rely on''). For Italian, a language-specific type was defined, namely inherently clitic verbs (e.g., ``prendersela''---eng. ``to be angry'').

The dataset used in the PARSEME shared task edition 1.2 contains 14 languages, including German (DE), Basque (EU), Greek (EL), French (FR), Irish (GA), Hebrew (HE), Hindi (HI), Italian (IT), Polish (PL), Brazilian Portuguese (PT), Romanian (RO), Swedish (SV), Turkish (TR), and Chinese (ZH). The number of tokens ranges from 35 k tokens (HI) to 1015 k tokens (RO), while the number of annotated VMWEs ranges from 662 (GA) to 9164 (ZH). The dataset split was made to ensure a higher number of unseen VMWEs in the dev (100 unseen VMWEs with respect to the train set) and test (300 unseen VMWEs with respect to the train + dev files) sets. More statistics regarding the PARSEME 1.2 dataset are depicted in Table \ref{tab:corpus_stats}.

In addition to the annotation with VMWEs, the multilingual PARSEME corpus is also tokenized, morphologically, and syntactically annotated, mostly with UDPipe \cite{udpipe:2017}. Thus, the syntactic analysis follows the principles of Universal Dependencies
 (\url{https://universaldependencies.org/} last accessed on 21 April 2023) \cite{ud}. 

\begin{table}[H]
    \centering
    \caption{The statistics of PARSEME 1.2: number of sentences (\#Sent.), of tokens (\#Tok.), and the sentence average length (Len.) on each of the three splits: training, validation, and test.}
    \label{tab:corpus_stats}
    
    \begin{tabularx}{\textwidth}{CccCCcCCcC}
         \toprule
         \multirow{ 2}{*}{\textbf{Lang.}} & \multicolumn{3}{c}{\textbf{Training}} & \multicolumn{3}{c}{\textbf{Validation}} & \multicolumn{3}{c}{\textbf{Test}} \\
          & \textbf{\#Sent.} & \textbf{\#Tok.} & \textbf{Len.} & \textbf{\#Sent.} & \textbf{\#Tok.} & \textbf{Len.} & \textbf{\#Sent.} & \textbf{\#Tok.} & \textbf{Len.} \\
         \midrule
         DE & 6.5 k & 126.8 k & 19.3 & 602 & 11.7 k & 19.5 & 1.8 k & 34.9 k & 19.1 \\
         EL & 17.7 k & 479.6 k & 27.0 & 909 & 23.9 k & 26.3 & 2.8 k & 75.4 k & 26.7 \\
         EU & 4.4 k & 61.8 k & 13.9 & 1.4 k & 20.5 k & 14.4 & 5.3 k & 75.4 k & 14.2 \\
         FR & 14.3 k & 360.0 k & 25.0 & 1.5 k & 39.5 k & 25.1 & 5.0 k & 126.4 k & 25.2 \\
         GA & 257 & 6.2 k & 24.2 & 322 & 7.0 k & 21.8 & 1.1 k & 25.9 k & 23.1\\
         HE & 14.1 k & 286.2 k & 20.2 & 1.2 k & 25.3 k & 20.2 & 3.7 k & 76.8 k & 20.2 \\
         HI & 282 & 5.7 k & 20.4 & 289 & 6.2 k & 21.7 & 1.1 k & 23.3 k & 21.0 \\
         IT & 10.6 k & 282.0 k & 27.4 & 1.2 k & 32.6 k & 27.1 & 3.8 k & 106.0 k & 27.3 \\
         PL & 17.7 k & 298.4 k & 16.8 & 1.4 k & 23.9 k & 16.8 & 4.3 k & 73.7 k & 16.7 \\
         PT & 23.9 k & 542.4 k & 22.6 & 1.9 k & 43.6 k & 22.1 & 6.2 k & 142.3 k & 22.8\\
         RO & 10.9 k & 195.7 k & 17.9 & 7.7 k & 134.3 k & 17.4 & 38.0 k & 685.5 k & 18.0 \\
         SV & 1.6 k & 24.9 k & 15.5 & 596 & 8.8 k & 14.9 & 2.1 k & 31.6 k & 15.0 \\
         TR & 17.9 k & 267.5 k & 14.9 & 1.0 k & 15.9 k & 15.0 & 3.3 k & 48.7 k & 14.7 \\
         ZH & 35.3 k & 575.5 k & 16.2 & 1.1 k & 18.2 k & 16.0 & 3.4 k & 55.7 k & 16.0\\
         \midrule
         Total & 175.7 k & 3512.7 k & 20.1 & 29.3 k & 522.2 k & 19.8 k & 81.9 k & 1581.6 k & 20.0 \\
         \bottomrule
    \end{tabularx}

\end{table}

\subsection{Fine-Tuning}

We followed the fine-tuning methodology employed by MTLB-STRUCT (the corresponding configuration files for each language are available at \url{https://github.com/shivaat/MTLB-STRUCT/tree/master/code/configs} last accessed on 21 April 2023) with the tree conditional random fields \cite{bradley2010learning} disabled. Thus, we trained our models for 10 epochs using a batch size of 32 and the Adam optimizer \cite{kingma2014adam}  with a learning rate of 3 $\times$ 10\textsuperscript{$-$5}. We set the maximum input sequence length to 150, the scaling parameter $k$, used in the gradient approximation of the lateral inhibition Heaviside function, to 10, which was empirically shown to create a good enough surrogate gradient \cite{pais-2022-racai}, and the hyperparameter $\lambda$ to $0.01$ in the DANN algorithm for scaling the reversed gradient. We did not employ k-fold cross-validation in our experiments, and we measured the model performance in terms of precision, recall, and F1-score {at the token level using the following equations:}
\begin{equation}
    {\textrm{Precision} = \frac{TP}{TP + FP}} \\
\end{equation}

\begin{equation}
    {\textrm{Recall} = \frac{TP}{TP + FN}} \\
\end{equation}

\begin{equation}
    {\textrm{F1-score} = \frac{2 \cdot Precision \cdot Recall}{Precision + Recall}} \\
\end{equation}
where $TP$ is the number of true positives, $FP$ is the number of false positives, and $FN$ is the number of false negatives. As suggested by the PARSEME 1.2 competition evaluation methodology (\url{https://www.davidsbatista.net/blog/2018/05/09/Named_Entity_Evaluation/} last accessed on 21 April 2023), we compute the strict variant of the F1-score. Thus, we consider the predicted label of a group of tokens as true positive only if it perfectly matches the ground truth \cite{sang2003introduction}.

\section{Results}
\label{sec:res}

The results of our evaluation for both monolingual and multilingual training, with and without lateral inhibition and adversarial training, for all the 14 languages, are displayed in Table \ref{tab:results}. We improved the performance of MTLB-STRUCT, the best overall system according to the competition benchmark (\url{https://multiword.sourceforge.net/PHITE.php?sitesig=CONF&page=CONF_02_MWE-LEX_2020___lb__COLING__rb__&subpage=CONF_40_Shared_Task} last accessed on 21 April 2023), on 11 out of 14 languages for global MWE prediction (the three remaining languages are German, Italian, and Romanian) and on 12 out of 14~languages for unseen MWE prediction (the two remaining languages are German and Greek). Out of all the cases where our methods underperformed, the only high difference was obtained in the German language, our best system being behind the MTLB-STRUCT system by approximately ~3.43\% on global MWE prediction and approximately ~6.57\% on unseen MWE prediction. We believe that this is due to the employment of the German BERT
 (\url{https://huggingface.co/bert-base-german-cased} last accessed on 21 April 2023) by the MTLB-STRUCT team, while we still used the mBERT model for \mbox{this language.}

\textls[-25]{For the global MWE prediction, we managed to improve the performance in  \mbox{11 languages,}} the highest F1-score was obtained by the monolingual training once (i.e., Chinese), by the simple multilingual training three times (i.e., Greek, Irish, and Turkish), by the multilingual training with lateral inhibition three times (i.e., French, Hebrew, and Polish), by the multilingual adversarial training once (i.e., Basque), and by the multilingual adversarial training with the lateral inhibition three times (i.e., Hindi, Portuguese, and Swedish). On the other hand, for the unseen MWE prediction, we managed to achieve better results in 12 languages. The simple multilingual training obtained the highest F1-score only once (i.e., Swedish), the multilingual training with the lateral inhibition three times (i.e., French, Turkish, and Chinese), the multilingual adversarial training five times (i.e., Irish, Hebrew, Hindi, Polish, and Romanian), and the multilingual adversarial training with lateral inhibition three times (i.e., Basque, Italian, and Portuguese).

\begin{table}[H]    
        \caption{The results obtained by the monolingual and multilingual training, together with the results obtained by the best system of the PARSEME 1.2 competition, MTLB-STRUCT. LI is the lateral inhibition component, while  Adv is the domain adaptation technique for cross-lingual MWE identification. {We measure the precision (P), recall (R), and F1-score (F1) for each global and unseen MWE identification experiment. The best results in each language are highlighted in bold.}}
        \label{tab:results}

\begin{adjustwidth}{-\extralength}{0cm}
%\centering %% If there is a figure in wide page, please release command \centering

  \begin{tabularx}{\fulllength}{CcCCCCCC}
     
        \toprule

        \multirow{ 2.1}{*}{\textbf{Language}} & \multirow{2.2}{*}{\textbf{Method}} &  \multicolumn{3}{c}{\textbf{Global MWE-Based}} & \multicolumn{3}{c}{\textbf{Unseen MWE-Based}} \\
        & & \textbf{P} & \textbf{R} & \textbf{F1} & \textbf{P} & \textbf{R} & \textbf{F1} \\
        \midrule
        
        \multirow{6}{*}{DE} & MTLB-STRUCT \cite{taslimipoor2020mtlb} & 77.11 & \textbf{75.24} & \textbf{76.17} & \textbf{49.17} & \textbf{49.50} & \textbf{49.34} \\
         & Monolingual & 74.26 & 72.82 & 73.53 & 40.35 & 41.79 & 41.06 \\
         & Multilingual & \textbf{77.26} & 68.47 & 72.60 & 37.85 & 43.22 & 40.35 \\
         & Multilingual + LI & 69.07 & 66.38 & 67.70 & 39.15 & 43.85 & 41.37 \\
         & Multilingual + Adv & 69.00 & 68.33 & 68.66 & 39.18 & 45.11 & 41.94 \\
         & Multilingual + LI + Adv & 71.37 & 68.08 & 69.69 & 41.47 & 43.85 & 42.77 \\
         \midrule
         
         \multirow{6}{*}{EL} & MTLB-STRUCT \cite{taslimipoor2020mtlb} & 72.54 & 72.69 & 72.62 & 38.74 & \textbf{47.00} & \textbf{42.47} \\
         & Monolingual & 72.33 & \textbf{73.00} & 72.66 & 38.30 & 46.75 & 42.11 \\
         & Multilingual & \textbf{74.60} & 72.38 & \textbf{73.48} & \textbf{38.92} & 42.21 & 40.50 \\
         & Multilingual + LI & 72.52 & 72.90 & 72.71 & 37.90 & 45.78 & 41.47 \\
         & Multilingual + Adv & 73.23 & 72.18 & 72.70 & 38.81 & 44.48 & 41.45 \\
         & Multilingual + LI + Adv & 73.42 & 72.59 & 73.00 & 38.64 & 44.16 & 41.21 \\
         \bottomrule
    
\end{tabularx}
\end{adjustwidth}
    %\caption{Caption}
    %\label{tab:my_label}
\end{table}

\begin{table}[H]    \ContinuedFloat
      \caption{{\em Cont.}}
        \label{tab:results}

\begin{adjustwidth}{-\extralength}{0cm}
%\centering %% If there is a figure in wide page, please release command \centering

  \begin{tabularx}{\fulllength}{CcCCCCCC}
     
        \toprule

        \multirow{ 2}{*}{\textbf{Language}} & \multirow{2}{*}{\textbf{Method}} &  \multicolumn{3}{c}{\textbf{Global MWE-Based}} & \multicolumn{3}{c}{\textbf{Unseen MWE-Based}} \\
        & & \textbf{P} & \textbf{R} & \textbf{F1} & \textbf{P} & \textbf{R} & \textbf{F1} \\
        \midrule
        
%        \multirow{6}{*}{DE} & MTLB-STRUCT \cite{taslimipoor2020mtlb} & 77.11 & \textbf{75.24} & \textbf{76.17} & \textbf{49.17} & \textbf{49.50} & \textbf{49.34} \\
%         & Monolingual & 74.26 & 72.82 & 73.53 & 40.35 & 41.79 & 41.06 \\
%         & Multilingual & \textbf{77.26} & 68.47 & 72.60 & 37.85 & 43.22 & 40.35 \\
%         & Multilingual + LI & 69.07 & 66.38 & 67.70 & 39.15 & 43.85 & 41.37 \\
%         & Multilingual + Adv & 69.00 & 68.33 & 68.66 & 39.18 & 45.11 & 41.94 \\
%         & Multilingual + LI + Adv & 71.37 & 68.08 & 69.69 & 41.47 & 43.85 & 42.77 \\
%         \midrule
%         
%         \multirow{6}{*}{EL} & MTLB-STRUCT \cite{taslimipoor2020mtlb} & 72.54 & 72.69 & 72.62 & 38.74 & \textbf{47.00} & \textbf{42.47} \\
%         & Monolingual & 72.33 & \textbf{73.00} & 72.66 & 38.30 & 46.75 & 42.11 \\
%         & Multilingual & \textbf{74.60} & 72.38 & \textbf{73.48} & \textbf{38.92} & 42.21 & 40.50 \\
%         & Multilingual + LI & 72.52 & 72.90 & 72.71 & 37.90 & 45.78 & 41.47 \\
%         & Multilingual + Adv & 73.23 & 72.18 & 72.70 & 38.81 & 44.48 & 41.45 \\
%         & Multilingual + LI + Adv & 73.42 & 72.59 & 73.00 & 38.64 & 44.16 & 41.21 \\
%         \midrule
%         
         \multirow{6}{*}{EU} & MTLB-STRUCT \cite{taslimipoor2020mtlb} & 80.72 & 79.36 & 80.03 & 28.12 & 44.33 & 34.41 \\
         & Monolingual & 81.61 & \textbf{80.40} & 81.00 & 34.94 & \textbf{49.29} & 40.89 \\
         & Multilingual & \textbf{86.49} & 77.03 & \textbf{81.49} & 33.32 & 45.04 & 39.17 \\
         & Multilingual + LI & 84.07 & 78.66 & 81.28 & 37.38 & 44.48 & 40.62 \\
         & Multilingual + Adv & 82.77 & 78.71 & 80.69 & 36.46 & 48.44 & 41.61 \\
         & Multilingual + LI + Adv & 84.80 & 78.42 & 81.48 & \textbf{39.71} & 46.46 & \textbf{42.82} \\
         \midrule
         
         \multirow{6}{*}{FR} & MTLB-STRUCT \cite{taslimipoor2020mtlb} & 80.04 & 78.81 & 79.42 & 39.20 & 46.00 & 42.33 \\
         & Monolingual & 79.84 & \textbf{79.54} & 79.69 & 38.89 & 44.87 & 41.67 \\
         & Multilingual & 81.80 & 77.04 & 79.35 & 43.17 & 44.55 & 43.85 \\
         & Multilingual + LI & \textbf{81.85} & 78.96 & \textbf{80.37} & \textbf{45.48} & \textbf{48.40} & \textbf{46.89} \\
         & Multilingual + Adv & 80.12 & 78.59 & 79.35 & 41.60 & \textbf{48.40} & 44.74 \\
         & Multilingual + LI + Adv & 80.47 & 78.22 & 79.33 & 40.87 & 45.19 & 42.92 \\
         \midrule
         
         \multirow{6}{*}{GA} & MTLB-STRUCT \cite{taslimipoor2020mtlb} & 37.72 & 25.00 & 30.07 & 23.08 & 16.94 & 19.54 \\
         & Monolingual & 33.67 & 23.17 & 27.45 & 24.02 & 17.28 & 20.10 \\
         & Multilingual & 54.91 & 34.63 & 42.48 & 45.91 & 28.61 & 35.25 \\
         & Multilingual + LI & 55.31 & 34.63 & 42.60 & 45.79 & 27.76 & 34.57 \\
         & Multilingual + Adv & \textbf{56.12} & \textbf{35.78} & \textbf{43.70} & \textbf{48.42} & \textbf{30.31} & \textbf{37.28} \\
         & Multilingual + LI + Adv & 55.72 & 34.63 & 42.72 & 45.79 & 27.76 & 34.57 \\
         \midrule
         
         \multirow{6}{*}{HE} & MTLB-STRUCT \cite{taslimipoor2020mtlb} & 56.20 & 42.35 & 48.30 & 25.53 & 15.89 & 19.59 \\
         & Monolingual & 54.09 & 40.76 & 46.49 & 26.02 & 15.94 & 19.77 \\
         & Multilingual & 61.38 & 40.76 & 48.98 & 34.76 & 17.81 & 23.55 \\
         & Multilingual + LI & \textbf{61.63} & 42.54 & \textbf{50.23} & 34.46 & 19.06 & 24.55 \\
         & Multilingual + Adv & 58.40 & 42.15 & 48.96 & \textbf{35.35} & \textbf{21.88} & \textbf{27.03} \\
         & Multilingual + LI + Adv & 59.89 & \textbf{42.74} & 49.88 & 34.92 & 20.62 & 25.93 \\
         \midrule

          \multirow{6}{*}{HI} & MTLB-STRUCT \cite{taslimipoor2020mtlb} & 72.25 & \textbf{75.04} & 73.62 & 48.75 & 58.33 & 53.11 \\
         & Monolingual & 66.53 & 70.28 & 68.35 & 49.35 & 61.35 & 54.70 \\
         & Multilingual & \textbf{77.78} & 71.77 & 74.65 & \textbf{62.72} & 58.65 & 60.61 \\
         & Multilingual + LI & 77.08 & 68.95 & 72.78 & 61.83 & 56.49 & 59.04 \\
         & Multilingual + Adv & 75.46 & 73.11 & 74.26 & 60.95 & \textbf{62.43} & \textbf{61.68} \\
         & Multilingual + LI + Adv & 75.53 & 73.85 & \textbf{74.68} & 60.31 & \textbf{62.43} & 61.35 \\
         \midrule

         \multirow{6}{*}{IT} & MTLB-STRUCT \cite{taslimipoor2020mtlb} & 67.68 & \textbf{60.27} & \textbf{63.76} & 20.23 & 21.33 & 20.81 \\
         & Monolingual & 64.53 & 59.59 & 61.96 & 20.81 & \textbf{24.06} & 22.32 \\
         & Multilingual & 69.37 & 56.40 & 62.21 & 22.22 & 19.38 & 20.70 \\
         & Multilingual + LI & \textbf{71.27} & 56.01 & 62.72 & 23.02 & 20.12 & 21.28 \\
         & Multilingual + Adv & 65.65 & 58.33 & 61.78 & 20.83 & 21.88 & 21.43 \\
         & Multilingual + LI + Adv & 69.18 & 57.85 & 63.01 & \textbf{25.51} & 23.44 & \textbf{24.43} \\
         \midrule

         \multirow{6}{*}{PL} & MTLB-STRUCT \cite{taslimipoor2020mtlb} & 82.94 & 79.18 & 81.02 & 38.46 & 41.53 & 39.94 \\
         & Monolingual & 81.89 & 79.33 & 80.85 & 38.30 & 41.99 & 40.06 \\
         & Multilingual & 84.02 & 77.03 & 80.37 & 40.34 & 37.50 & 38.87 \\
         & Multilingual + LI & \textbf{85.14} & 79.26 & \textbf{82.09} & \textbf{44.48} & 41.33 & 42.84 \\
         & Multilingual + Adv & 82.55 & \textbf{79.85} & 81.18 & 40.75 & \textbf{45.19} & \textbf{42.86} \\
         & Multilingual + LI + Adv & 83.19 & 78.74 & 80.90 & 41.01 & 41.67 & 41.34 \\
         \midrule

         \multirow{6}{*}{PT} & MTLB-STRUCT \cite{taslimipoor2020mtlb} & 73.93 & 72.76 & 73.34 & 30.54 & 41.33 & 35.13 \\
         & Monolingual & 74.81 & 70.94 & 73.01 & 33.81 & 39.05 & 35.98 \\
         & Multilingual & 75.93 & 70.94 & 73.35 & 34.06 & 39.18 & 36.44 \\
         & Multilingual + LI & \textbf{77.15} & 71.89 & 74.43 & \textbf{35.61} & 39.18 & 37.31 \\
         & Multilingual + Adv & 73.36 & 73.48 & 73.42 & 30.33 & 40.13 & 34.55 \\
         & Multilingual + LI + Adv & 75.51 & \textbf{73.53} & \textbf{74.49} & 33.76 & \textbf{41.78} & \textbf{37.36} \\
         \bottomrule
    
\end{tabularx}
\end{adjustwidth}
    %\caption{Caption}
    %\label{tab:my_label}
\end{table}

\begin{table}[H]    \ContinuedFloat
      \caption{{\em Cont.}}
        \label{tab:results}

\begin{adjustwidth}{-\extralength}{0cm}
%\centering %% If there is a figure in wide page, please release command \centering

  \begin{tabularx}{\fulllength}{CcCCCCCC}
     
        \toprule

        \multirow{ 2}{*}{\textbf{Language}} & \multirow{2}{*}{\textbf{Method}} &  \multicolumn{3}{c}{\textbf{Global MWE-Based}} & \multicolumn{3}{c}{\textbf{Unseen MWE-Based}} \\
        & & \textbf{P} & \textbf{R} & \textbf{F1} & \textbf{P} & \textbf{R} & \textbf{F1} \\
        \midrule
   \multirow{6}{*}{RO} & MTLB-STRUCT \cite{taslimipoor2020mtlb} & 89.88 & \textbf{91.05} & \textbf{90.46} & 28.84 & 41.47 & 34.02 \\
         & Monolingual & 90.39 & 90.11 & 90.25 & 46.82 & 51.09 & 48.86 \\
         & Multilingual & 91.34 & 88.46 & 89.88 & \textbf{49.90} & 48.12 & 48.99 \\
         & Multilingual + LI & 90.78 & 88.85 & 89.81 & 45.06 & 45.15 & 45.10 \\
         & Multilingual + Adv & 89.14 & 90.13 & 89.63 & 46.27 & \textbf{56.44} & \textbf{50.85} \\
         & Multilingual + LI + Adv & 89.95 & 88.78 & 89.36 & 45.44 & 50.30 & 47.74 \\
         \midrule
         
         \multirow{6}{*}{SV} & MTLB-STRUCT \cite{taslimipoor2020mtlb} & 69.59 & 73.68 & 71.58 & 35.57 & 53.00 & 42.57 \\
         & Monolingual & 73.01 & 73.68 & 73.34 & 44.32 & \textbf{54.62} & 48.93 \\
         & Multilingual & \textbf{78.92} & 70.79 & 74.63 & \textbf{50.78} & \textbf{54.62} & \textbf{52.63} \\
         & Multilingual + LI & 75.48 & 73.68 & 74.57 & 46.77 & 52.66 & 49.54\\
         & Multilingual + Adv & 75.42 & \textbf{74.41} & 74.91 & 46.70 & 53.50 & 49.87 \\
         & Multilingual + LI + Adv & 77.62 & 74.10 & \textbf{75.82} & 49.47 & 51.82 & 50.62 \\
         \midrule
         
         \multirow{6}{*}{TR} & MTLB-STRUCT \cite{taslimipoor2020mtlb} & 68.41 & 70.55 & 69.46 & 42.11 & 45.33 & 43.66 \\
         & Monolingual & 69.11 & 72.89 & 70.95 & 43.75 & 47.88 & 45.72 \\
         & Multilingual & 67.52 & \textbf{73.27} & \textbf{71.18} & 41.83 & 47.56 & 44.51 \\
         & Multilingual + LI & \textbf{69.92} & 72.28 & 71.08 & \textbf{47.94} & \textbf{49.19} & \textbf{48.55}\\
         & Multilingual + Adv & 68.41 & 70.37 & 69.38 & 43.54 & 47.23 & 45.31 \\
         & Multilingual + LI + Adv & 68.22 & 69.77 & 68.99 & 43.04 & 44.30 & 43.66 \\
         \midrule
         
         \multirow{6}{*}{ZH} & MTLB-STRUCT \cite{taslimipoor2020mtlb} & 68.56 & 70.74 & 69.63 & 58.97 & 53.67 & 56.20 \\
         & Monolingual & \textbf{72.33} & \textbf{72.88} & \textbf{72.60} & 59.74 & \textbf{58.03} & 58.87 \\
         & Multilingual & 72.03 & 71.32 & 71.67 & 62.30 & 55.87  & 58.91 \\
         & Multilingual + LI & 69.82 & 70.36 & 70.09 & 62.50 & 57.31 & \textbf{59.79}\\
         & Multilingual + Adv & 69.29 & 69.47 & 69.38 & 62.42 & 54.73 & 58.32 \\
         & Multilingual + LI + Adv & 70.64 & 68.58 & 69.59 & \textbf{65.41} & 54.73 & 59.59 \\
         \bottomrule
    
\end{tabularx}
\end{adjustwidth}
    %\caption{Caption}
    %\label{tab:my_label}
\end{table}

Also, the monolingual training has not achieved the highest F1-score for unseen MWE prediction for any language.~These findings are summarized in Table \ref{tab:highest_f1}.

\begin{table}[H]
    \centering
    \caption{The number of times we managed to obtain the highest F1-score with each system developed in this work for both global MWE (\#Highest Global MWE) and unseen MWE (\#Highest Unseen MWE) predictions.}
    \begin{tabularx}{\textwidth}{CCC}
        \toprule
        \multirow{2}{*}{\textbf{Method}} & \textbf{\#Highest} & \textbf{\#Highest} \\
         & \textbf{Global MWE} & \textbf{Unseen MWE} \\ 
         \midrule
         MTLB-STRUCT \cite{taslimipoor2020mtlb} & 3 & 2 \\
          \midrule
         Monolingual & 1 & 0 \\
         Multilingual & 3 & 1 \\
         Multilingual + LI & 3 & 3 \\
         Multilingual + ADV & 1 & 5 \\
         Multilingual + LI + ADV & 3 & 3 \\
         \midrule
         Total (ours) & 11 & 12 \\
         \bottomrule
    \end{tabularx}
    \label{tab:highest_f1}
\end{table}

We further compared the average scores across all languages obtained by our systems. In Table \ref{tab:avg_results}, we compared our results with the ones obtained by each system at the latest edition of the PARSEME competition (\textls[-25]{{\url{https://multiword.sourceforge.net/PHITE.php?sitesig=CONF&page=CONF_02_MWE-LEX_2020___lb__COLING__rb__&subpage=CONF_50_Shared_task_results}} last accessed on 21 April 2023}): MTLB-STRUCT \cite{taslimipoor2020mtlb}, Travis-multi/mono \cite{kurfali2020travis}, Seen2Unseen \cite{pasquer2020seen2unseen}, FipsCo \cite{corpus-1.2}, HMSid \cite{colson2020hmsid}, and MultiVitamin \cite{gombert2020multivitaminbooster}. For the global MWE identification, we outperformed the MTLB-STRUCT results with all the multilingual training experiments, the highest average F1-score being obtained by the simple multilingual training without lateral inhibition or adversarial training. It achieved an average F1-score of 71.37\%, an improvement of 1.23\% compared to the MTLB-STRUCT F1-score (i.e., 70.14\%). For unseen MWE identification, we improved the average results obtained by MTLB-STRUCT using all the methodologies employed in this work. The highest average F1-score was obtained by the multilingual adversarial training with 43.26\%, outperforming the MTLB-STRUCT system by 4.73\%.

\begin{table}[H]
    \centering
    \caption{The average precision (AP), recall (AR), and F1-scores (AF1) over all languages obtained by our systems are compared with the results obtained by each system at the PARSEME 1.2 competition on global and unseen MWE identification. We also depict the number of languages used to train each system (\#Lang). The best results are highlighted in bold.}
    \begin{tabularx}{\textwidth}{cCCCCCCC}
         \toprule
        \multirow{2}{*}{\textbf{Method}} & \multirow{2}{*}{\textbf{\#Lang.}} &  \multicolumn{3}{c}{\textbf{Global MWE-Based}} & \multicolumn{3}{c}{\textbf{Unseen MWE-Based}} \\
        & & \textbf{AP} & \textbf{AR} & \textbf{AF1} & \textbf{AP} & \textbf{AR} & \textbf{AF1} \\
        \midrule
        MTLB-STRUCT \cite{taslimipoor2020mtlb} & 14/14 & 71.26 & \textbf{69.05} & 70.14 & 36.24 & 41.12 & 38.53 \\
        TRAVIS-multi \cite{kurfali2020travis} & 13/14 & 60.65 & 57.62	& 59.10 & 28.11 & 33.29 & 30.48 \\
        TRAVIS-mono \cite{kurfali2020travis} & 10/14 & 49.50 & 43.48 & 46.34 & 24.33 & 28.01 & 26.04 \\
        Seen2Unseen \cite{pasquer2020seen2unseen} & 14/14 & 63.36 & 62.69 & 63.02 & 16.14 & 11.95 & 13.73 \\ 
        FipsCo \cite{corpus-1.2} & 3/14 & 11.69 & 8.75 & 10.01 & 4.31 & 5.21 & 4.72 \\ 
        HMSid \cite{colson2020hmsid} & 1/14 & 4.56 & 4.85 & 4.70 & 1.98 & 3.81 & 2.61 \\ 
        MultiVitaminBooster \cite{gombert2020multivitaminbooster} & 7/14 & 0.19 & 0.09 & 0.12 & 0.05 & 0.07 & 0.06 \\ 
        \midrule
        Monolingual & 14/14 & 70.60 & 68.52 & 69.54 & 38.52 & 42.42 & 40.38 \\
        Multilingual & 14/14 & \textbf{75.23} & 67.88 & \textbf{71.37} & 42.72 & 41.60 & 42.15 \\
        Multilingual + LI & 14/14 & 74.36 & 68.24 & 71.17 & \textbf{43.48} & 42.20 & 42.78 \\
        Multilingual + Adv & 14/14 & 72.78 & 68.92 & 70.80 & 42.26 & \textbf{44.30} & \textbf{43.26} \\
        Multilingual + LI + Adv & 14/14 & 73.96 & 68.56 & 71.16 & 43.24 & 42.75 & 43.00 \\
        \bottomrule
    \end{tabularx}
    \label{tab:avg_results}
\end{table}

\vspace{-12pt}
\section{Discussion}
\label{sec:discussion}

\begin{figure}[H]
    \includegraphics[width=0.84\linewidth]{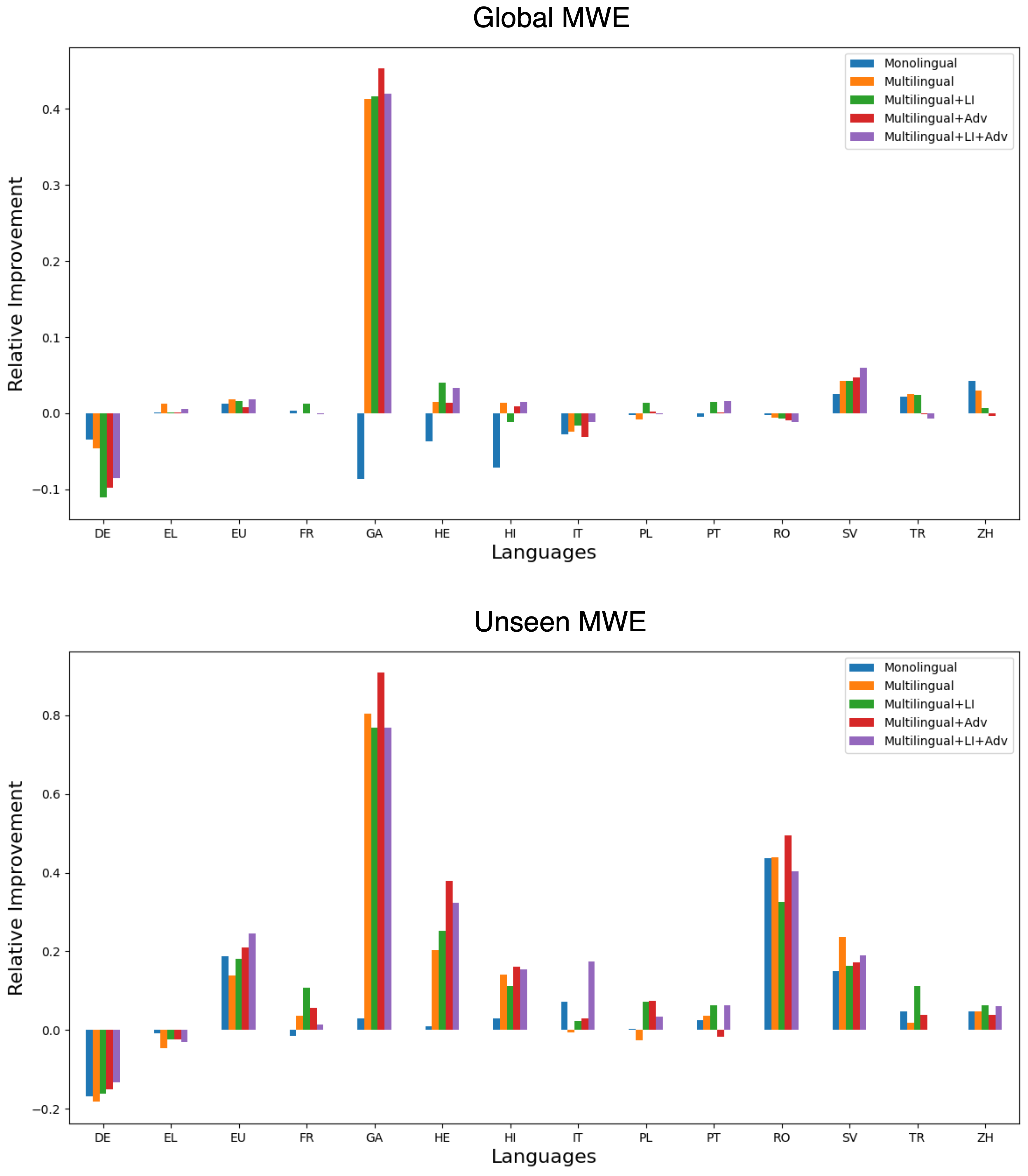}
    \caption{Improvements brought by our methodologies (i.e., Monolingual, Multilingual, Multilingual+LI, Multilingual+Adv, and Multilingual+LI+Adv) on global (\textbf{top}) and unseen (\textbf{bottom}) MWE prediction compared to the results of MTLB-STRUCT, the best system in the PARSEME shared task edition 1.2.}
    \label{fig:improve_global}
\end{figure}

According to our experiments, the average MWE identification performance can be improved by approaching this problem using a multilingual NLP system, as described in this work. An interesting perspective of our results on this task is how much improvement we brought compared to the PARSEME 1.2 competition's best system. These results are shown at the top of Figure \ref{fig:improve_global} for global MWE prediction and at its bottom for unseen MWE prediction. In general, the most significant relative improvements were achieved in the Irish language by employing multilingual training that, combined with adversarial training, boosted the performance by 45.32\% for the global MWE prediction and by 90.78\% for the unseen MWE prediction. On the other hand, for the same language, by using the monolingual training, we decrease the system's performance on global MWE prediction by 8.71\% and slightly increase it by 2.86\% on unseen MWE prediction. We believe that these improvements in Irish were due to the benefits brought by the multilingual training since this language contained the least amount of training sentences (i.e., 257 sentences), and it has been shown in previous research that superior results are obtained when such fine-tuning mechanisms are employed \cite{eisenschlos2019multifit}. However, the Hindi language also contains a small number of training samples (i.e., 282 sentences), but our multilingual training results are worse when compared to Irish. We assume that this is the outcome of the language inequalities that appeared in the mBERT pre-training data \cite{wu2020all} and the linguistic isolation of Hindi since there are no other related languages in the fine-tuning data \cite{dhamecha2021role}.

The second highest improvements for global MWE prediction were achieved in the Swedish language with 2.45\% for the monolingual training, 4.26\% for the multilingual training, 4.17\% for the multilingual training with the lateral inhibition, 4.65\% for the multilingual adversarial training, and 5.92\% for the multilingual adversarial training with lateral inhibition. We observe a relatively high difference between the first and the second place, but we believe again that this is due to the small number of sentences for Irish compared to Swedish. On the other hand, the results for unseen MWE prediction outline that the second highest improvements were attained in Romanian with 43.62\% for the monolingual training, 44.00\% for the multilingual training, 32.56\% for the multilingual training with lateral inhibition, 49.47\% for the multilingual adversarial training, and 40.32\% for the multilingual adversarial training with lateral inhibition. In addition, the improvements are more uniform on the unseen MWE prediction than the global one.

\section{Conclusions and Future Work}
\label{sec:concl}

Failure to identify MWEs can lead to misinterpretation of text and errors in NLP tasks, making this an important area of research. In this paper, we analyzed the performance of MWE identification in a multilingual setting, training the mBERT model on the combined PARSEME 1.2 corpus using all the 14 languages found in its composition. In addition, to boost the performance of our system, we employed lateral inhibition and language adversarial training in our methodology, intending to create embeddings that are as language-independent as possible. Our evaluation results highlighted that through this approach, we managed to improve the results obtained by MTLB-STRUCT, the best system of the PARSEME 1.2 competition, on 11 out of 14 languages for global MWE identification and 12 out of 14 for unseen MWE identification. Thus, with the highest average F1-scores of 71.37\% for global MWE identification and 43.26\% for unseen MWE identification, we class ourselves over MTLB-STRUCT by 1.23\% for the former task and by 4.73\% for the latter.

Possible future work directions involve analyzing how the language-independent features produced by mBERT are when lateral inhibition and adversarial training are involved, together with an analysis of more models that produce multilingual embeddings, such as XLM or XLM-R. In addition, we intend to analyze these two methodologies, with possible extensions, for multilingual training beyond MWE identification, targeting tasks, such as language generation or named entity recognition. Finally, since the languages in the PARSEME 1.2 dataset may share similar linguistic properties, we would like to explore how language groups improve each other's performance in the multilingual scenario.

%%%%%%%%%%%%%%%%%%%%%%%%%%%%%%%%%%%%%%%%%%
% \authorcontributions{Conceptualization, X.X. and Y.Y.; methodology, X.X.; software, X.X.; validation, X.X., Y.Y. and Z.Z.; formal analysis, X.X.; investigation, X.X.; resources, X.X.; data curation, X.X.; writing---original draft preparation, X.X.; writing---review and editing, X.X.; visualization, X.X.; supervision, X.X.; project administration, X.X.; funding acquisition, Y.Y. All authors have read and agreed to the published version of the manuscript.}

\vspace{6pt}
\authorcontributions{Conceptualization, A.-M.A., V.B.M., V.P. and D.-C.C.; methodology, A.-M.A. and V.P.; software, A.-M.A.; validation, A.-M.A., V.B.M., D.-C.C. and Ș.T.-M.; formal analysis, A.-M.A.; investigation, A.-M.A., V.B.M. and D.-C.C.; resources, A.-M.A. and V.B.M.; data curation, A.-M.A.; writing---original draft preparation, A.-M.A., V.B.M. and V.P.; writing---review and editing, A.-M.A., V.B.M., D.-C.C. and Ș.T.-M.; visualization, A.-M.A.; supervision, D.-C.C. and Ș.T.-M.; project administration, D.-C.C.; funding acquisition, D.-C.C. All authors have read and agreed to the published version of the manuscript.}

\funding{This research has been funded by the University Politehnica of Bucharest through the PubArt program.}

\dataavailability{The PARSEME 1.2 dataset used in this work has been open-sourced by the competition organizers and is available for public usage at \url{https://lindat.mff.cuni.cz/repository/xmlui/handle/11234/1-3367} (last
accessed on 21 April 2023).} 

%\acknowledgments{In this section, you can acknowledge any support given which is not covered by the author contribution or funding sections. This may include administrative and technical support, or donations in kind (e.g., materials used for experiments).}

\conflictsofinterest{The authors declare no conflict of interest.} 

%%%%%%%%%%%%%%%%%%%%%%%%%%%%%%%%%%%%%%%%%%
\begin{adjustwidth}{-\extralength}{0cm}

\reftitle{References}

\PublishersNote{}
\end{adjustwidth}
\end{document}